\documentclass[sigconf]{acmart}

\usepackage{booktabs}
\usepackage[ruled,vlined,linesnumbered]{algorithm2e}
\usepackage{array}
\usepackage{multirow}
\usepackage{epsfig}

\begin{document}
\title{Automated Problem Identification: Regression vs Classification via Evolutionary Deep Networks}

   \author{Emmanuel Dufourq \textsuperscript{1, 2}, Bruce A. Bassett \textsuperscript{1, 2, 3}\\
\small{\textsuperscript{1} Department of Mathematics and Applied Mathematics, University of Cape Town, Rondebosch, 7701, South Africa \\
\textsuperscript{2} African Institute for Mathematical Sciences, 6 Melrose Road, Muizenberg, 7945, Cape Town, South Africa\\
\textsuperscript{3} South African Astronomical Observatory, Observatory, Cape Town, 7925, South Africa\\ 
emails: $^{1}$ edufourq@gmail.com, $^{2}$ bruce.a.bassett@gmail.com}}
    \affiliation{ 
      \institution{}
    }

\begin{abstract}

Regression or classification? This is perhaps the most basic question faced when tackling a new supervised learning problem. We present an Evolutionary Deep Learning (EDL) algorithm that automatically solves this by identifying the question type with high accuracy, along with a proposed deep architecture. Typically, a significant amount of human insight and preparation is required prior to executing machine learning algorithms. For example, when creating deep neural networks, the number of parameters must be selected in advance and furthermore, a lot of these choices are made based upon pre-existing knowledge of the data such as the use of a categorical cross entropy loss function. Humans are able to study a dataset and decide whether it represents a classification or a regression problem, and consequently make decisions which will be applied to the execution of the neural network. We propose the Automated Problem Identification (API) algorithm, which uses an evolutionary algorithm interface to TensorFlow to manipulate a deep neural network to decide if a dataset represents a classification or a regression problem. We test API on 16 different classification, regression and sentiment analysis datasets with up to 10,000 features and up to 17,000 unique target values. API achieves an average accuracy of $96.3\%$ in identifying the problem type without hardcoding any insights about the general characteristics of regression or classification problems. For example, API successfully identifies classification problems even with 1000 target values. Furthermore, the algorithm recommends which loss function to use and also recommends a neural network architecture. Our work is therefore a step towards fully automated machine learning.

\end{abstract}

\maketitle

\newenvironment{conditions}
  {\par\vspace{\abovedisplayskip}\noindent\begin{tabular}{>{$}l<{$} @{${}={}$} l}}
  {\end{tabular}\par\vspace{\belowdisplayskip}}

\section{Introduction}

As the performance of machine learning algorithms has skyrocketed over recent years the often unspoken relationship between the human data scientist and the machines they run has evolved significantly. 
A great deal of work has been put into new state-of-the-art methods, and researchers are constantly optimising the various aspects of machine learning algorithms. Such efforts include proposing algorithms for optimising hyperparameters and network architectures \cite{real:2017:large} and the latest trends show increasing emphasis on algorithms that require less human intervention. Consider the automatic statistician project \footnote{https://www.automaticstatistician.com/index/} which aims at removing the data scientist from the process of understanding data by using Bayesian model selection. Real {\em et al}. \cite{real:2017:large} propose an evolutionary algorithm for optimising image classification neural networks which requires no human intervention in creating the networks. Similarly, Zoph and Le \cite{zoph:2016:neural} use recurrent neural networks along with reinforcement learning  in order to achieve a similar goal. It is clear from these research efforts that this is a trend that will continue, driven both by potential industrial profits to compensate for shortages of expensive data scientists and by the general goal of Artificial General Intelligence (AGI).

Nevertheless, for most current machine learning algorithms, there is a considerable amount of human intervention which must be performed prior to the final execution of the algorithm. For example, setting the number of parameters, preprocessing the data, deciding on the loss function and interpreting the results, to name a few.  Another example, and perhaps the first of the steps in the data science process, is problem identification: {\em "does a supervised set of data correspond to a classification or regression problem?"} Understanding which type of the two problems a given dataset represents is a step in the direction of automated machine learning research and is the subject of this study.

\begin{figure*}[!h] 
  \centering
          \includegraphics[width=0.75\textwidth]{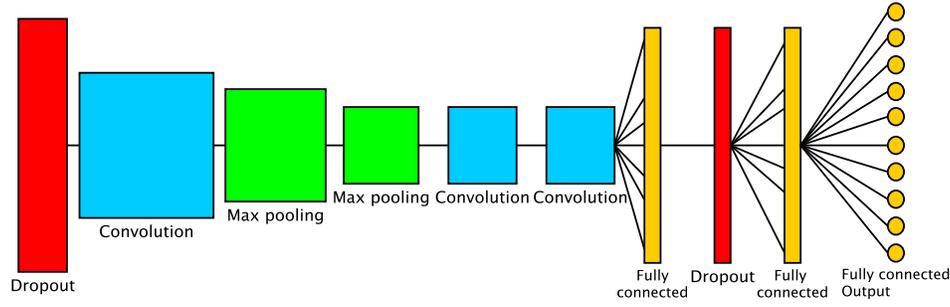}
  \caption{Each chromosome contains four genes of which one gene represents a network architecture. The figure illustrates an example of a network architecture generated by an API chromosome (which was obtained at the end of an execution of the API algorithm). The input dataset was CIFAR-10 -- an image classification dataset. The chromosome recommended that the last layer should have 10 units and that these should use the sigmoid activation function. Furthermore, the chromosome recommended using the categorical cross entropy loss function, and consequently, correctly determined that the dataset was a classification problem.}\label{fig-configuration}
\end{figure*}

Classification problems typically represent a set of problems whereby the goal is to create a predictive model that can discriminate between various known classes. CIFAR-10 and MNIST are examples of classification datasets where the goals are to identify the correct label (airplane, automobile, bird, cat etc... and digits respectively) for each image. For regression problems, the predictive output is continuous (as opposed to discrete in the case of classification). An example of a regression dataset is the Boston housing price regression dataset for which the goal is to predict the median value of the houses. 

In the context of deep learning \cite{lecun:2015:deep}, when presented with a dataset, typically one will verify whether the data represents a classification or a regression problem, and then will decide on the loss function and network layers accordingly. For the CIFAR-10 image dataset, one might consider using convolutional, dropout and fully connected layers; and for the Boston housing price dataset one might use fully connected and dropout layers. Furthermore, a decision should be made with regards to which loss function (or equivalently, figure of merit) to use. For CIFAR-10 one might use categorical cross entropy, and use the mean squared error loss function for the Boston housing case. As researchers in machine learning, in most cases, these decisions can be made with relative ease. For a machine, on the other hand, this decision is non-trivial and current machine learning algorithms do not automatically decide if a given dataset is a classification or a regression problem; nor do they recommend a loss function.

In this study, a genetic algorithm (GA) harnessed to a dynamic and flexible deep learning framework is proposed for the automated identification of problems. We call this the Automated Problem Identification (API) algorithm and show that it can successfully determine if a dataset is a classification or a regression one; and furthermore, recommend whether to use categorical cross entropy or mean-squared error. Additionally, API will recommend which layers (e.g. convolutional or fully connected) -- from a known set --  to use, either as the final architecture or as the input to further optimisation. Figure \ref{fig-configuration} illustrates an example of a network which was produced by a chromosome when the CIFAR-10 dataset was input into API. The resulting architecture is very similar to one that a human might use for the problem. 

This paper is organized as follows: Section 2 describes GAs. Section 3 describes the API chromosome which is used to determine if a dataset is a classification or a regression optimisation problem. Section 4 provides the details for the proposed API algorithm. The experimental setup is presented in Section 5 and Section 6 discusses the results. We conclude in Section 7 and discus our future work.

\section{Genetic Algorithm}\label{sec: gas}

A Genetic Algorithm (GA) \cite{Goldberg:1989:GeneticAlgorithms} is a biologically inspired evolutionary algorithm \cite{Eiben:2003:IntroTo}. GAs mimic the way that species fight for survival and reproduce in nature. A GA makes use of a population of chromosomes to solve an optimisation problem. Each chromosome encodes a potential solution to the problem. Over time the chromosomes undergo many modifications, known as genetic operators, in order to traverse the search space. A fitness function is used to determines how good a chromosome is at solving the optimisation problem. Each generation parent chromosomes are selected and genetic operators are applied to those parents to create offspring which then constitute the new chromosome -- and parent -- population. The new population is evaluated for fitness and the process is repeated; as illustrated in Algorithm \ref{algo:ga-algo}.

\begin{algorithm}

\SetKwData{Gen}{generation}
\SetKwData{Genmax}{generation\_max}
\SetKwInOut{Input}{input}

\Input{\Genmax : maximum number of GA generations}

	\Begin{
	
	Create an initial population of chromosomes.
	
	Evaluate the initial population.
	
	\Gen $\leftarrow 0$.
	
	\While{\Gen $\leq$ \Genmax}{
	
		\Gen $\leftarrow \Gen+1.$
	
		Select the parents.
		
		Perform the genetic operators.
		
		Replace the current population with the new offspring created in step 8.

		Evaluate the current population.
	
	}

\Return{The best chromosome.}

}
\caption{Genetic algorithm} 
\label{algo:ga-algo} 
\end{algorithm}

\section{Proposed API Chromosome} \label{chromosome}

In this section and the following subsections, we describe the API chromosome along with a description about each of the genes within the chromosome. In this study, the word layer refers to the layers in deep neural network architectures. Each chromosome is made up of four genes, namely, the neural network loss function, the number of units in the last layer of the neural network, the activation function used in the last layer and the configuration of the layers (configurations are explained in section \ref{configuration}). A chromosome thus encodes an entire deep neural network architecture and an associated loss function. 

\begin{figure}[!ht]
  \centering
          \includegraphics[width=0.40\textwidth]{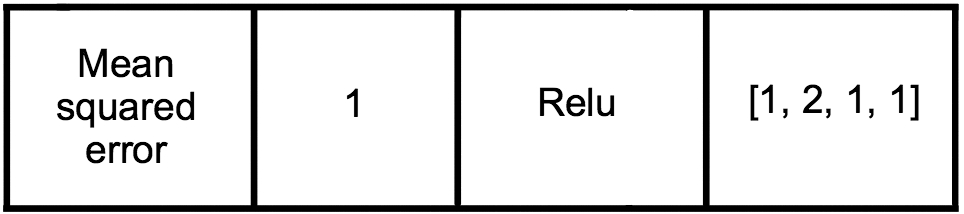}
  \caption{Example of an API chromosome which encodes the mean squared error loss function, 1 unit in the last layer of the network which has the relu activation function. The architecture of the network, denoted as [1, 2, 1, 1] represents a fully connected layer, followed by dropout and two fully connected layers. The configurations are explained in section \ref{configuration}.}
\label{fig:chromosome}
\end{figure}

Figure \ref{fig:chromosome} illustrates an example of an API chromosome that encodes a neural network architecture with the following layers:  fully connected, dropout and two fully connected layers. Furthermore, the chromosome will apply the mean squared error loss function (during the training of the neural network) and the last layer has 1 unit of which the activation function is a rectified linear unit. The following subsections provide additional details about the four genes.

We chose to use GAs since the number of genes can easily be modified in order to encode additional complexity and to easily handle the discrete nature of the parameters being chosen, since API searches through a space of network architectures in addition to other parameters. We can increase the complexity of the chromosomes by including more parameters, as we discuss in section \ref{conclusion}.

\subsection{Loss Function}

This gene represents the loss function that will be used when training the network and it takes on two possible values: Mean Squared Error (MSE) and Categorical Cross Entropy (CCE) loss. Let $y_{i}$ denote the target label for sample $i$, $\bar{y}_{i}$ denote the model's predicted output for sample $i$ and $N$ denote the number of training samples. The mean-squared error used in this study is presented in equation 1. 

\begin{equation}
\frac{1} {N}{\sum\limits_{i = 1}^N {(y_{i} - \bar{y}_{i} } })^{2} 
\end{equation}

For example, let $y = [2.2 \; 5.5 \; 0.2]$ and assume that some network $N_1$ predicts $\bar{y} = [2.1 \; 5.0 \; 0.2]$ then $MSE_{N_{1}} \approx 0.09$. Similarly, assume that another network $N_2$ predicts $\bar{y} = [0.0 \; 2.5 \; 3.2]$ then $MSE_{N_{2}} \approx 7.61$. Network $N_1$ is preferred since $0.09 < 7.61$. When using the MSE, the objective of an optimisation algorithm will be to minimise the MSE value to reduce the distance between the correct values and the model's predicted values. 

The categorical cross entropy used in this study is presented in equation 2. 

\begin{equation}
{-\sum\limits_{i = 1}^N {y_{i} \ln \bar{y}_{i} } } 
\end{equation}

When using this loss function the objective is to maximise the CCE in such a way to make the network predictions are as similar to the labels as possible. In this case, the target labels will be represented as a vector (often one-hot encoded vectors) and the network predictions will also be in a vector of the same length. For example, let $y_{i} = [0 \; 1 \; 0]$ and assume that some network $N_1$ predicts $\bar{y}_{i} = [0.1 \; 0.8 \; 0.1]$ then CCE for sample $i$ is  $-\ln(0.8)$. Similarly, assume that another network $N_2$ predicts $\bar{y}_{i} = [0.7 \; 0.1 \; 0.2]$ then for sample $i$, CCE $= -\ln(0.1)$. Network $N_2$ is preferred since $-\ln(0.1) > -\ln(0.7)$.

\subsection{Number of Units in Last Layer}

The second gene denotes the number of units in the last layer in the network. There are two possible values for this gene: one or $U$, where $U$ denotes the number of unique values in $Y$ ($Y$ represents the target values for a dataset). Formally, $U = \vert S \vert$, where $S = \{y_i\}_{i\in\{1,\ldots,N\}}$ and $N$ is the number of samples. For example, assume that for some dataset $Y = (0, 1, 0, 2, 3, 2, 3, 0, 2, 3, 1, 1) $ then $U=4$ since there are 4 unique values in the targets.

\subsection{Last Layer Function}

This gene takes on four possible values and denotes which activation function will be used in the last layer in the network. The possible values are: \{linear, relu, sigmoid, softmax\}. Here `relu' refers to rectified linear units. Given some input $x$ to a layer, the equations for each of the activation functions are presented in equations 3 to 6 respectively. 

\begin{equation}
{f(x) = x} 
\end{equation}

\begin{equation}
{f(x) = \max (x, 0)} 
\end{equation}

\begin{equation}
{f(x) = \frac{1}{(1+e^{-x})}} 
\end{equation}

\begin{equation}
{f(x) = \frac{e^{x_j}} {\sum\limits_{i = 1}^{D} {e^{x_i} } } }
\end{equation} where

\begin{conditions}
 D     &  dimension of $x$\\
 j     & 1, ..., D \\   
\end{conditions}

\subsection{Configuration of Layers}\label{configuration}

Each chromosome has a gene which corresponds to the architecture of the network which we define as the configuration. The configuration represents the exact sequence of the network layers and is stored in a list. The first element in the configuration represents the first layer and the last element represents the last layer. There are four possible values which each element in the configuration can take, namely: convolution \cite{lecun:1989:generalization}, fully connected, dropout \cite{Srivastava:2014:Dropout} and max pooling \cite{Zhou:1988:Computation}. Here, convolution refers to two-dimensional convolution. We add dropout to the list of possible configuration values even though dropout is not a layer.

The size of the configurations is randomly selected between 5 and 15. The configurations are initialised randomly during the initial population generation and modified during the mutation operator; these are explained in sections \ref{initpop} and \ref{mutation} respectively. Each of the layers are mapped to an integer value, i.e. convolution is mapped to 0, fully connected to 1, dropout to 2 and max pooling to 3. Each chromosome has exactly one configuration.

We provide the following example to illustrate the configurations. Let the configuration for a chromosome be: [2, 0, 3, 3, 0, 0, 1, 2, 1, 1]; figure \ref{fig-configuration} illustrates this network. The network is comprised of several convolution and max pooling layers followed by fully connected and dropout layers.

\subsection{Chromosome Fitness Evaluation}

GAs make use of a fitness function to evaluate how good a chromosome is at solving an optimisation problem. In our case, we designed a fitness function to discriminate between classification and regression problems. When the proposed system commences, it splits the dataset into two subsets, the features, $X$ and the labels $Y$. The labels are then converted into their corresponding one-hot encoded values. For example, if a label has a value of 2 and the unique $Y$ values are \{0, 1, 2\} then the one-hot encoded value of `2' is [0 0 1]. The system retains both $Y$ and the one-hot encoded $Y$ values. The dataset is split such that 50\% of the data is in the training set and the remaining in the validation set.

Each chromosome is evaluated as follows. The chromosome's loss function is used to train the neural network on the training data. If the chromosome's loss function was categorical cross entropy, then the one-hot encoded $Y$ values are used during training. However, if the loss function was mean squared error, then the $Y$ values are used during training.

The validation loss is recorded during the optimisation of the neural network across the epochs. Let the validation loss be $V_0, V_1, ..., V_e$ where $e$ denotes the total number of epochs and $V_i$ denote the validation loss for epoch $i$. We define the change in validation as follows, $\delta_{val} = average(V_1, V_2, ..., V_e) - V_0.$ Finally, we define the ratio in validation drop, $R$, as $R = \frac{\delta_{val}} {V_0}.$ Thus, for each chromosome, after the optimisation of the neural network has taken place, we compute $R$, and if $R>0$ then this implies that the network has not done any learning since the validation loss increased. Furthermore, if $R = 0$ then once again the network has not managed to learn anything since the validation loss has remained constant over the epochs. Finally, if $R < 0$ we conclude that given the drop in validation loss, that the network has managed to learn. 

The model then predicts the output values on the validation data. The predictions and the validation target values are compared using mean squared error. The loss obtained on the validation data using categorical cross entropy will be different to the loss computed using mean squared error. We chose to use the mean squared error to be consistent with the comparisons. In the case whereby the network has not learnt anything we penalise the chromosome with a fitness of infinity. However, in the case whereby the network has learnt, i.e. $R < 0$, then we assign the computed validation mean squared error as the fitness of the chromosome. The objective of the API algorithm is thus to minimise the fitness of each chromosome by rewarding networks that learn and have a small mean squared error on the validation set. The lower the fitness value the better a chromosome performed. 

\begin{figure}[!ht]
  \centering
          \includegraphics[width=0.30\textwidth]{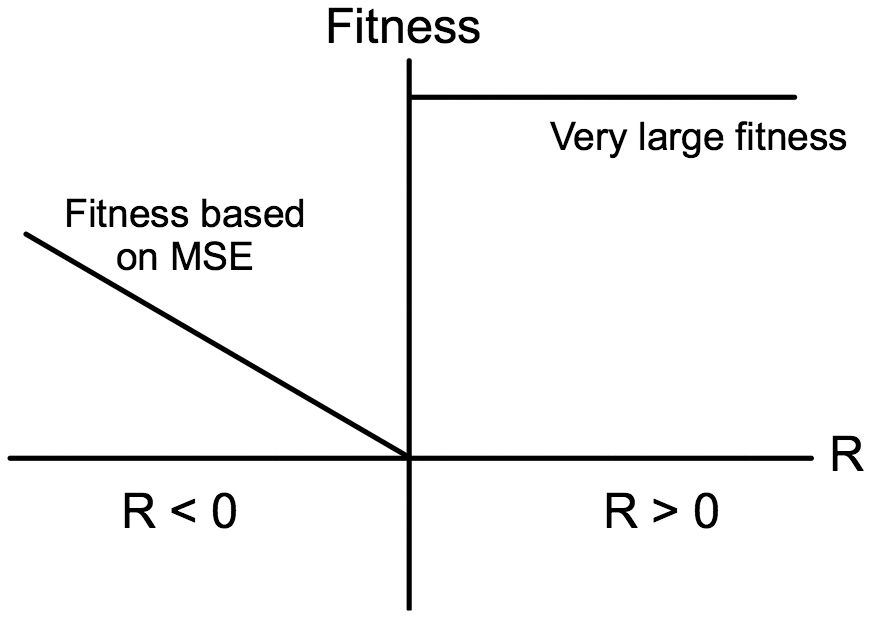}
  \caption{During the training of each neural network on the validation data we record the validation loss. We then determine whether or not the network has learnt. If the network has not learnt ($R \ge 0$) then we penalise the chromosome with a very large fitness. However, if the network was able to learn ($R < 0$) then we assign the mean squared error as the fitness value.}
\label{fitnessplot}
\end{figure}

Figure \ref{fitnessplot} illustrates a plot which explains the fitness of a chromosome. The plot is separated in two where $R = 0$. From the plot, it is observed that when $R \ge 0$ then the fitness is set to a very large value. When $R < 0$ then the value of the fitness corresponds to the mean squared error whereby a smaller value is better. For the sake of the example, a straight line was drawn for $R < 0$ to illustrate that a smaller mean squared error results in a better chromosome fitness. 

Since the chromosomes are randomly generated, it is possible that they represent invalid networks for particular features and labels on a given dataset. For example, assume that, for some chromosome, the number of units in the last layer is 1 and the categorical cross entropy loss function is used. Given the previous description in this subsection, the one-hot encoded $Y$ values should be used during training. However, in the example, the number of outputs is 1 and thus a one-hot encoded vector cannot be compared to a single value. To illustrate with another example, consider a chromosome that tries to use convolutional layers on a feature based regression dataset - this is, of course, not feasible. Invalid chromosomes such as this are penalised with a fitness of infinity. Section \ref{decision} describes how the chromosome makes the discrimination between regression and classification.

\section{The API Algorithm} \label{algorithm-main}

The following subsections explain how each aspect of the GA has been adapted to determine if a given dataset is a classification or regression problem. Furthermore, the algorithm recommends the following upon termination: the loss function which should be used in order to enable the training of a neural network, the number of units and type of activation function in the last layer and finally, a simple network architecture is also recommended. The API algorithm performs optimisation in two phases, namely in optimising the GA population, and since each chromosome represents a neural network, optimisation is performed when training the networks. 

\subsection{Initial Population Generation} \label{initpop}

The initial population size is set to the same value as the user-defined population size. Suppose the population size is \textit{n}, then \textit{n} chromosomes are created during the initial population generation. Each chromosome has a fixed length of 4 genes (discussed in section \ref{chromosome}). During the creation of a chromosome, each gene is randomly created based on the values each gene can take on. The pseudocode for creating a chromosome is presented in algorithm \ref{algo:create-chromosome}. The initial fitness of each chromosome is set to infinity.

\begin{algorithm}

\SetKwInOut{Input}{input}

	\Begin{
	
	Initialise an empty chromosome.

	Set the loss function to either categorical cross entropy or mean squared error.

	Set the number of units in the last layer to either one or $U$.

	Set the activation function in the last layer to either linear, sigmoid, softmax or relu.

	Create a random configuration.

}
\caption{Creating a chromosome.} 
\label{algo:create-chromosome} 
\end{algorithm}

\subsection{Parent Selection}\label{sub:parentSelection}

Parent selection methods are used to obtain parents from the current population of chromosomes. These parents are used by the genetic operators to create offspring. A single parent is obtained when the parent selection method is executed. Once a chromosome has been chosen to be a parent, the selection method can  select that particular chromosome again. Three common parent selection methods are fitness-proportionate, rank and tournament selection \cite{Blickle:1996:AComparison}. For this study, tournament selection was used given that it was shown to be a successful method by Zhong et al. \cite{Zhong:2005:ComparisonOfPerformance}.

Algorithm \ref{algo:tournament-selection} presents the pseudocode for the tournament selection. This selection method has one user-defined parameter, namely, the tournament size. Let \textit{k} be the tournament size. Tournament selection randomly selects \textit{k} chromosomes from the current GA population, and compares the fitness of each of the \textit{k} chromosomes. The chromosome with the lowest fitness is returned as the parent chromosome. If a tie occurs, then a random chromosome is selected to break the tie.

\begin{algorithm}

\SetKwData{Size}{size}
\SetKwData{Current}{current\_best}
\SetKwData{Chromosome}{random\_chromosome}
\SetKwInOut{Input}{input}
\SetKwInOut{Output}{output}

\Input{\Size : size of the tournament}
\Output{The best chromosome which will be used as a parent}

\Begin{

	\Current $\leftarrow$ null
	
	\For{$i\leftarrow 1$ \KwTo \Size}{
		
		\Chromosome $\leftarrow$ randomly select a chromosome from the population

		Evaluate \Chromosome

		\If{fitness of \Chromosome $<$ fitness of \Current}{

			\Current $\leftarrow$ \Chromosome

		}

	}

\Return{\Current}

}
\caption{Pseudocode for tournament selection.} 
\label{algo:tournament-selection} 
\end{algorithm}

\subsection{Genetic Operators}

Genetic operators are applied to parents to exchange genetic material between the parent chromosomes, and to consequently create novel offspring. The two most common genetic operators are mutation and crossover. Their implementation details for this study are described below.

\subsubsection{Mutation}\label{mutation}

\begin{figure}[!ht]
  \centering
          \includegraphics[width=0.40\textwidth]{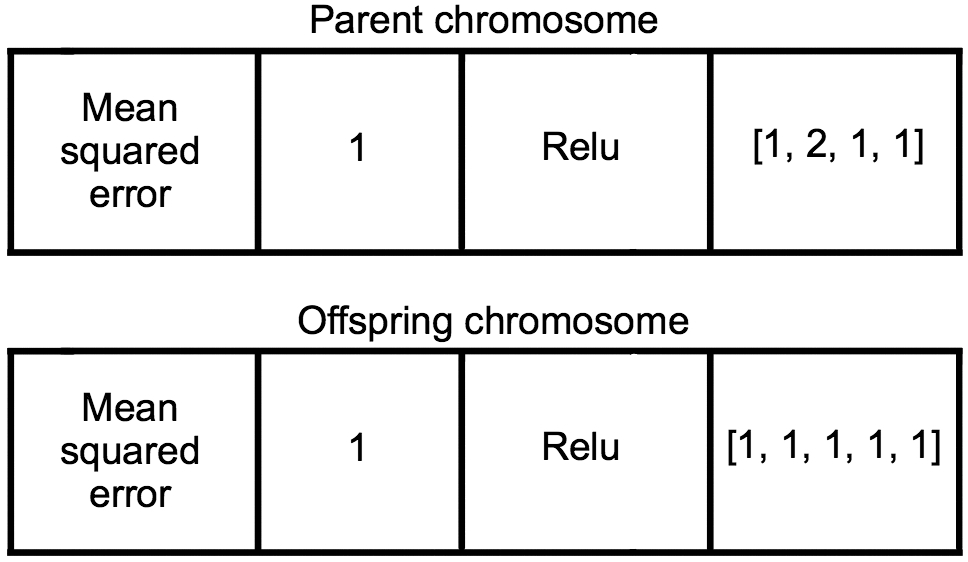}
  \caption{Example of the mutation operator being applied to a parent chromosome. The forth gene was selected for mutation and consequently a new configuration was generated for the offspring.}
\label{fig:mutation}
\end{figure}

The mutation genetic operator makes use of a single parent chromosome. During the execution of mutation, a gene is randomly selected and a new value for that gene is created. A user-defined parameter is associated with the mutation operator, namely the mutation application rate. Figure \ref{fig:mutation} illustrates the application of the mutation operator on a parent chromosome, and the resulting offspring is illustrated. From the example, the forth gene was selected for mutation and thus the forth gene within the parent was changed from a configuration of [1, 2, 1, 1] to [1, 1, 1, 1, 1] in the offspring.

\subsubsection{Crossover}

The crossover genetic operator exchanges genetic material between two parent chromosomes: $parent_{1}$ and $parent_{2}$, and consequently creates two offspring: \textit{offspring}$_{1}$ and \textit{offspring}$_{2}$. There are several variations of the crossover genetic operator, such as uniform, one-point and two-point crossover. 

The crossover method we implement randomly selects a position \textit{p} in the range $[0,n]$ --- where \textit{n} denotes the length of the chromosome --- within the parent chromosomes; the same position \textit{p} must be selected within the two parents. Two offspring are created, and all the genes except those at position \textit{p} are copied across to the corresponding offspring without modification. The genes are position \textit{p} are swapped, i.e., the gene in position \textit{p} from $parent_1$ is inserted into position \textit{p} in \textit{offspring}$_2$, and similarly, the gene in position \textit{p} from $parent_2$ is inserted into position \textit{p} in \textit{offspring}$_1$. 

\begin{figure}[!ht]
  \centering
          \includegraphics[width=0.48\textwidth]{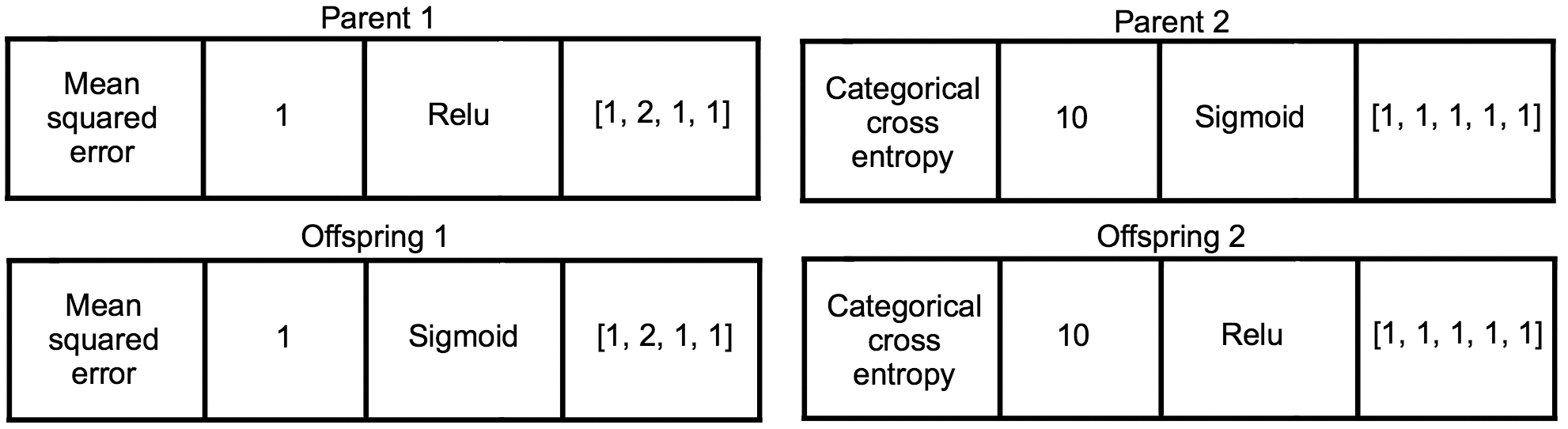}
  \caption{Example of the crossover operator being applied two parent chromosomes. The third gene from both of the parents were selected for crossover. As a result, the last activation functions were swapped between the parents. }
\label{fig:crossover}
\end{figure}

An example of the application of the crossover operator is presented in figure \ref{fig:crossover}. The figure shows two parent chromosomes. The crossover point was the third gene from each parent, i.e. the last activation function was swapped between the parents. The offspring show the result of the crossover.

\subsection{Algorithm Termination and Final Decision} \label{decision}

At the end of the generational loop, the best chromosome is output. The loss function in the best chromosome is then used to decide if the dataset was a classification or a regression problem. If the loss function was categorical cross entropy, then the problem was labelled as classification. However, if the loss function was mean squared error then the problem was labelled as regression.

\section{Experimental Setup} \label{experimentalsetup}

This section describes the experimental set up which was used to evaluate the performance of API. The algorithm was programmed in Python 3.6.0 and TensorFlow 1.1.0 \cite{tensorflow2015-whitepaper}. The algorithm was evaluated on a machine with an Intel Core i7-6700K CPU and 16GB RAM.

\subsection{Datasets}

Table \ref{datasets} presents the 16 datasets which were used in this study along with their characteristics and type. All of the datasets were obtained from the UCI machine learning repository \cite{Lichman:2013} except for CIFAR-10 and CIFAR-100 which were obtained from \cite{krizhevsky:2009:learning}, MNIST from \cite{LeCun:1998:Gradient}, CrowdFlower \footnote{https://www.crowdflower.com/data-for-everyone/}, Aloi \footnote{https://www.csie.ntu.edu.tw/~cjlin/libsvmtools/datasets/multiclass.html} and IMDB \footnote{https://keras.io/datasets/} were obtained externally. In this study, CrowdFlower represents the `emotion in text' dataset. The assumptions are that there are no missing values in each dataset and that categorical features are converted to corresponding numerical features using a one-hot encoding approach. Of course, it would be possible to implement an imputation method \cite{mcknight:2007:missing} to overcome datasets with missing values, however, this was not part of the scope of this study. The algorithm standardises each feature. Where possible, we used 1000 samples for training and 1000 for validation. Boston housing, for example, did not have that many samples. In this case, we simply split the dataset equally into two sets. We distinguish between data and image classification problems because in the former the data are typically resented by one-dimensional vectors; whereas, image classification datasets are commonly represented as three-dimensional arrays. API can adapt to the various input shapes without human intervention.

\begin{table}
\begin{centering}
\begin{tabular}{cccc}
\hline 
\textbf{Dataset} & \textbf{Features} & \textbf{Unique Targets} & \textbf{Type}\tabularnewline
\hline 
Aloi & 128 & 1000 & D\tabularnewline
Isolet5 & 617 & 26 & D\tabularnewline
Letter Recognition & 16 & 26 & D\tabularnewline
Sensorless Drive & 48 & 11 & D\tabularnewline
Year Prediction & 90 & 89 & D\tabularnewline
Boston Housing & 13 & 506 & R\tabularnewline
CCPP & 4 & 4837 & R\tabularnewline
Concrete Comp & 15 & 1030 & R\tabularnewline
Forest Fires & 29 & 17380 & R\tabularnewline
Pysiocochemical & 9 & 15903 & R\tabularnewline
Relative CT Slice & 384 & 2001 & R\tabularnewline
CIFAR-10 & 3072 & 10 & IC\tabularnewline
CIFAR-100 & 3072 & 100 & IC\tabularnewline
MNIST & 784 & 10 & IC\tabularnewline
CrowdFlower & 1000 & 13 & SA\tabularnewline
IMDB & 10000 & 2 & SA\tabularnewline
\hline 
\end{tabular}
\par\end{centering}
\caption{The 16 datasets used in this study. We used datasets from four problem
domains with various characteristics and are denoted as follows: `D'
represents data classification, `R' for regression, `IC' for image
classification and `SA' for sentiment analysis. The sentiment analysis
datasets were considered as classification problems. The unique targets
refers to the unique number of outputs in the target values for each
dataset. For example, for CIFAR-10 has 10 unique target classes, whereas
Relative CT Slice has 15903 unique target values. For CrowdFlower
and IMDB we used a bags of words approach in order to generate word
embeddings.} \label{datasets}
\end{table}

\subsection{Experimental Parameters}

The GA and neural network parameters used in this study are presented in tables \ref{ga-parameters} and \ref{nn-parameters} respectively. These parameters were obtained by preliminary runs of the algorithm. The purpose of this study was to evolve chromosomes that could determine whether a given dataset was classification or regression in addition to several other outputs. Certain variables had to remain fixed in order to evolve the chromosomes. Each parameter in table \ref{nn-parameters} was set to a fixed value.

\begin{table} 
\begin{centering}
\begin{tabular}{cc}
\hline 
\textbf{Parameter} & \textbf{Value}\tabularnewline
\hline 
Crossover rate & 70\%\tabularnewline
Mutation rate & 30\%\tabularnewline
Number of generations & 10\tabularnewline
Tournament size & 5\tabularnewline
Population size & 50\tabularnewline
\hline 
\end{tabular}
\par\end{centering}
\caption{The GA parameters used in this study. Preliminary experiments revealed
that we did not need to use a large population size or a large number
of generations to evolve accurate solutions.} \label{ga-parameters}
\end{table}

\section{Results and Discussion}\label{results}

\begin{figure}[!ht]
  \centering
          \includegraphics[width=0.48\textwidth]{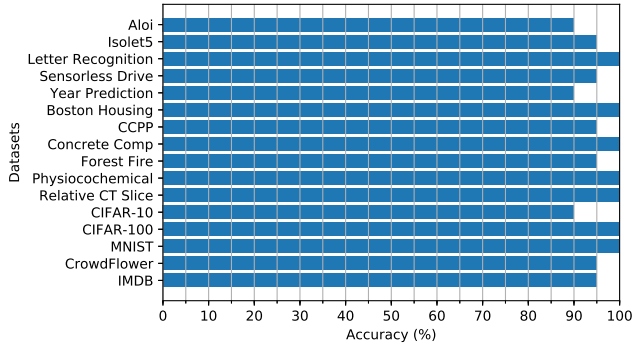}
  \caption{Accuracy (\%) results obtained by API on the various datasets. For each dataset, 20 runs of the algorithm were executed. The lowest accuracy was 90\% and API achieved 100\% accuracy on 7 datasets. The average accuracy across the datasets was 96.3\%.} \label{figplot}
\end{figure}

The results obtained by API are presented and discussed in this section. The Aloi dataset was included in the experiments because one could hypothesise that if a dataset has a large number of targets then it is a regression dataset. For this reason, we included Aloi as it has a much larger number of classes in comparison to the other classification datasets. The accuracy results achieved by API on the 16 datasets across the 20 runs are presented in figure \ref{figplot}. When discriminating between regression or classification problems, API obtained an average accuracy of 96.3\%, the lowest accuracy was 90\% which was obtained on 3 datasets and the highest accuracy was 100\% which was achieved on 7 datasets.

Table \ref{table-incorrect} presents the number of times, out of 20 runs, that API incorrectly classified each dataset. There were 3 datasets for which API incorrectly classified two runs,  this represents an accuracy of 90\%. There was no dataset for which the performance across the runs was less than 90\%.

In the case of the two misclassifications for the CIFAR-10 dataset, the fitness for chromosomes having the mean squared error and the categorical cross entropy loss function were very close. It happened to be that, for that particular run, the former had a slightly lower fitness. In the second case, the population was rapidly dominated by chromosomes having the mean squared error loss function as the generational loop progressed. A similar observation was made for the other incorrectly classified runs. Two possible ways of overcoming this issue would be to re-introduce genetic diversity into the population by randomly initialising a number of chromosomes across the generations. This would thus allow chromosomes containing both types of loss functions to be present in the population. Alternatively, increasing the tournament size could allow for weaker chromosomes to remain in the population which could in turn preserve the balance between the chromosomes containing both loss functions. 

\begin{table} 
\begin{centering}
\begin{tabular}{cc}
\hline 
\textbf{Parameter} & \textbf{Value}\tabularnewline
\hline 
Number of epochs & 5\tabularnewline
Weight initialisation - mean & 0.0\tabularnewline
Weight initialisation - standard deviation & 0.01\tabularnewline
Number of units in all layers except last & 100\tabularnewline
Activation functions in all layers except last & relu\tabularnewline
Number of filters in each convolution layer & 10\tabularnewline
Convolution filter size & 2x2\tabularnewline
Convolution strides & 1\tabularnewline
Max pooling size & 2x2\tabularnewline
Max pooling stride & 1\tabularnewline
Dropout keep probability & 0.8\tabularnewline
Learning rate & 0.001\tabularnewline
Optimiser & Adam \cite{kingma2014adam} \tabularnewline
Batch size & 2048\tabularnewline
\hline 
\end{tabular}
\par\end{centering}
\caption{The neural network parameters used in this study. When training a neural network contained in a 
chromosome each of the parameters listed in this table were applied.} \label{nn-parameters}
\end{table}

\begin{table}
\begin{centering}
\begin{tabular}{cc>{\centering}p{4cm}}
\hline 
\textbf{Dataset} & \textbf{Type} & \textbf{Number of Incorrectly Classified Runs}\tabularnewline
\hline 
Aloi & D & 2\tabularnewline
Isolet5 & D & 1\tabularnewline
Letter Recognition & D & 0\tabularnewline
Sensorless Drive & D & 1\tabularnewline
Year Prediction & D & 2\tabularnewline
Boston Housing & R & 0\tabularnewline
CCPP & R & 1\tabularnewline
Concrete Comp & R & 0\tabularnewline
Forest Fire & R & 1\tabularnewline
Physiocochemical & R & 0\tabularnewline
Relative CT Slice & R & 0\tabularnewline
CIFAR-10 & IC & 2\tabularnewline
CIFAR-100 & IC & 0\tabularnewline
MNIST & IC & 0\tabularnewline
CrowdFlower & SA & 1\tabularnewline
IMDB & SA & 1\tabularnewline
\hline 
\end{tabular}
\par\end{centering}
\caption{The table presents the number of runs for which the algorithm incorrectly
classified each dataset. The objective of API was to discriminate
between regression and classification datasets. For each dataset we
performed 20 API runs. A perfect accuracy of 100\% was achieved
on 7 datasets. For the types, `D' represents data classification,
`R' for regression, `IC' for image classification and `SA' for sentiment
analysis.} \label{table-incorrect}
\end{table}

Appendix A presents, for each dataset, an example chromosome that was evolved. These chromosomes were randomly selected from each of the 20 runs. The networks varied in size from 5 to 15 layers, however, in most cases the networks were deep. The architecture generated for the image classification problems are more complex than the ones generated for the other problems. In particular, the evolved chromosome for the CIFAR-10 dataset was of interest because the configuration resembles an architecture that a human might generated when creating a deep neural network for image classification. For instance, consider AlexNet \cite{krizhevsky:2009:learning}, which is made up of a series of convolutional and max pooling layers towards the start of the network, and ends with three fully connected layers. In a similar way, the chromosome's architecture which is presented in the appendix has a similar structure of convolutions and max pooling layers followed by fully connected and dropout layers. 

Some of the other chromosomes in the other runs for CIFAR-10 evolved similar architectures, but this was not always the case. For example, in one particular run, the evolved architecture was: [0, 0, 0, 2, 2, 2, 2, 0, 0, 1]. In this case, the architecture was primarily made up of dropout and convolutional layers -- there was only one fully connected layer. For certain runs, the evolved architectures were made up of deep networks containing only fully connected layers. For example, from the appendix, consider the chromosome presented for the Sensorless Drive dataset; the architecture was [1, 1, 1, 1, 1, 1, 1, 1, 1, 1, 1, 1, 1, 1, 1, 1].

The number of epochs used throughout the optimisation of the neural networks was small. It would thus be of interest to extend this study in order to investigate the architectures which would be generated by using a larger number of epochs. One drawback of API is the computational effort required to obtain the results. It would be of interest to further decrease the population size to determine to which extent it can be reduced whilst retaining its current accuracy in discriminating between classification and regression problems.

\section{Conclusion}\label{conclusion}

In our study, we present the Automated Problem Identification (API) algorithm, a genetic algorithm coupled to deep networks to automatically determining whether a dataset represents a regression or classification problem. While great effort has been put into improving and proposing new machine learning algorithms, typically the practitioner must still decide on the loss function, neural network architecture, number of units in each layer and select appropriate activation functions prior to the execution of the neural network. We propose API with the goal of moving towards general artificial intelligence and automated machine learning that requires little to no human intervention.

API was applied to 20 times each to 16 different datasets drawn from varied problem domains and data characteristics. We find that API correctly identified the problem type with an average accuracy of 96.3\% running only a single CPU. Furthermore, API was able to recommend whether to use mean squared error or categorical cross entropy, a suitable number of units in the last layer together with the activation function, and furthermore, recommend a network architecture. Despite not being the primary focus of this study, the proposed algorithm generated interesting and relevant deep architectures.

We have already begun working on the next phase of this research which is to develop an algorithm which can optimise the entire pipeline for creating deep neural networks; whereby, the goal is simply to provide the algorithm with a dataset  (without specifying if the problem is a classification or regression problem) and in return, get a deep neural network which can produce competitive results. This would completely remove the human from the pipeline. It would be of interest to determine if the evolved networks could outperform those created by humans. It is clear, with the efforts of various researchers that the machine learning community should steer towards algorithms which are completely automated requiring no human intervention.

\appendix

\section{Examples of API chromosomes}\label{appendixA}

Here we illustrate examples of API chromosomes which were evolved on the various datasets. The dataset name is provided along with the problem type. For the last activation function, `MSE' denotes mean squared error, and `CCE' denotes categorical cross entropy. For the configurations, convolution is mapped to 0, fully connected to 1, dropout to 2 and max pooling to 3. In each example the chromosome was able to correctly classify the dataset.

\begin{itemize}
\item \textbf{Dataset:} Aloi -- Classification \\ \textbf{Chromosome:} Units: 1000, Loss: CCE, Activation: linear, Configuration:  [2, 1, 1, 2, 1]\\
\item \textbf{Dataset:} Isolet5 -- Classification \\ \textbf{Chromosome:} Units: 1, Loss: MSE, Activation: softmax, Configuration: [1, 1, 1, 1, 1, 1, 1, 1, 1, 1, 1] \\
\item \textbf{Dataset:} Letter Recognition -- Classification  \\ \textbf{Chromosome:} Units: 26, Loss: CCE, Activation: sigmoid, Configuration: [1, 2, 2, 1, 2, 2, 1, 1, 2, 1, 1] \\
\item \textbf{Dataset:} Sensorless Drive -- Classification  \\ \textbf{Chromosome:} Units: 11, Loss: CCE, Activation: relu, Configuration:  [1, 1, 1, 1, 1, 1, 1, 1, 1, 1, 1, 1, 1, 1, 1, 1]\\
\item \textbf{Dataset:} Year Prediction -- Classification  \\ \textbf{Chromosome:} Units: 64, Loss: CCE, Activation: sotmax, Configuration: [1, 2, 2, 2, 1, 2, 2, 1, 2, 2, 1] \\
\item \textbf{Dataset:} Boston Housing -- Regression \\ \textbf{Chromosome:} Units: 1, Loss: MSE, Activation: softmax, Configuration:  [2, 1, 1, 2, 1, 1, 1, 1, 2, 1] \\
\item \textbf{Dataset:} CCPP -- Regression \\ \textbf{Chromosome:} Units: 1, Loss: MSE, Activation: softmax, Configuration:  [1, 2, 2, 2, 1, 1, 1, 1, 2, 2, 1]\\
\item \textbf{Dataset:} Concrete Comp -- Regression \\ \textbf{Chromosome:} Units: 1, Loss: MSE, Activation: softmax, Configuration:  [1, 1, 1, 1, 1, 1, 1, 1, 1, 1, 1, 1, 1, 1, 1] \\
\item \textbf{Dataset:} Forest Fire --Regression \\ \textbf{Chromosome:} Units: 1, Loss: MSE, Activation: softmax, Configuration:  [1, 2, 2, 2, 1, 1, 2, 2, 1, 2, 2, 2, 1, 1, 2, 1] \\
\item \textbf{Dataset:} Pysiocochemical -- Regression \\ \textbf{Chromosome:} Units: 1, Loss: MSE, Activation: softmax, Configuration: [1, 1, 1, 2, 2, 1, 1, 2, 1, 1, 1, 1, 1, 2, 1] \\
\item \textbf{Dataset:} Relative CT Slice -- Regression \\ \textbf{Chromosome:} Units: 1, Loss: MSE, Activation: softmax, Configuration: [1, 2, 1, 1, 2, 1, 1] \\
\item \textbf{Dataset:} CIFAR-10  -- Image classification  \\ \textbf{Chromosome:} Units: 10, Loss: CCE, Activation: linear, Configuration:  [3, 3, 0, 0, 2, 3, 3, 0, 0, 0, 1] \\
\item \textbf{Dataset:} CIFAR-100  -- Image classification  \\ \textbf{Chromosome:} Units: 100, Loss: CCE , Activation: sigmoid, Configuration:  [2, 0, 3, 3, 0, 0, 1, 2, 1, 1, 1]\\
\item \textbf{Dataset:} MNIST  -- Image classification  \\ \textbf{Chromosome:} Units: 10, Loss: CCE, Activation: relu, Configuration: [2, 0, 2, 0, 3, 0, 1] \\
\item \textbf{Dataset:} CrowdFlower -- Sentiment analysis  \\ \textbf{Chromosome:} Units: 13, Loss: CCE, Activation: sigmoid, Configuration:  [1, 2, 1, 1, 1, 2, 2, 1, 2, 2, 1, 1, 1, 2, 1]\\
\item \textbf{Dataset:} IMDB -- Sentiment analysis  \\ \textbf{Chromosome:} Units: 2, Loss: CCE, Activation: softmax, Configuration:  [2, 1, 2, 1, 2, 1]\\
\end{itemize}

\begin{acks}
The financial assistance of the National Research Foundation (NRF) towards this research is hereby acknowledged. Opinions expressed and conclusions arrived at, are those of the author and are not necessarily to be attributed to the NRF. The authors would like to thank Ethan Roberts, Shankar Agarwal and Arun K. Aniyan for their feedback and comments.
\end{acks}

\bibliographystyle{unsrt}

\bibliography{References} 

\end{document}